\documentclass[letterpaper, 10 pt, conference]{ieeeconf}  

\IEEEoverridecommandlockouts                              

\usepackage{graphics} 
\usepackage{amsmath} 
\usepackage{amssymb}  
\usepackage{graphicx}
\usepackage{comment}
\usepackage{multirow}
\usepackage{moresize}
\usepackage{soul}
\usepackage{color}
\usepackage{url}
\usepackage{float}
\usepackage{caption}
\usepackage{subcaption}

\title{\LARGE \bf Gust Estimation and Rejection with a Disturbance Observer for Proprioceptive Underwater Soft Morphing Wings}

\author{Tobias Cook$^{1}$, Leo Micklem$^{1}$, Huazhi Dong$^{2}$, Yunjie Yang$^{2}$, Michael Mistry$^{3}$, and Francesco Giorgio-Serchi$^{1}$
\thanks{*This work was not supported by any grant.} 
\thanks{$^{1}$Tobias Cook, Leo Micklem and Francesco Giorgio-Serchi are with the Institute of Integrated Micro and Nano Systems, School of Engineering, The University of Edinburgh, Edinburgh, UK. (Corresponding author: Francesco Giorgio-Serchi, {\tt\small f.giorgio-serchi@ed.ac.uk})}%
\thanks{$^{2}$Huazi Dong and Yunjie Yang are with Institute of Digiatal Communciation, The University of Edinburgh, Edinburgh, UK.
}%
\thanks{$^{3}$Michael Mistry is with the School of Informatics, The University of Edinburgh, Edinburgh, UK.
}
}

\begin{document}

\maketitle
\thispagestyle{empty}
\pagestyle{empty}

\begin{abstract}
Unmanned underwater vehicles are increasingly employed for maintenance and surveying tasks at sea, but their operation in shallow waters is often hindered by hydrodynamic disturbances such as waves, currents, and turbulence. These unsteady flows can induce rapid changes in direction and speed, compromising vehicle stability and manoeuvrability. Marine organisms contend with such conditions by combining proprioceptive feedback with flexible fins and tails to reject disturbances. Inspired by this strategy, we propose soft morphing wings endowed with proprioceptive sensing to mitigate environmental perturbations. The wing’s continuous deformation provides a natural means to infer dynamic disturbances: sudden changes in camber directly reflect variations in the oncoming flow. By interpreting this proprioceptive signal, a disturbance observer can reconstruct flow parameters in real time. To enable this, we develop and experimentally validate a dynamic model of a hydraulically actuated soft wing with controllable camber. We then show that curvature-based sensing allows accurate estimation of disturbances in the angle of attack. Finally, we demonstrate that a controller leveraging these proprioceptive estimates can reject disturbances in the lift response of the soft wing. By combining proprioceptive sensing with a disturbance observer, this technique mirrors biological strategies and provides a pathway for soft underwater vehicles to maintain stability in hazardous environments.
\end{abstract}

\section{Introduction}

The use of soft robots is becoming more widespread for a variety of applications. Due to their flexibility, they can be compliant with their surroundings, and bio-inspired designs offer the possibility of increased efficiency and manoeuvrability. As such, they have found uses as manipulators \cite{hughes2016soft}, grippers \cite{shintake2018soft}, and actuators for untethered robots \cite{rich2018untethered}. One possible area of application is as a propulsion or steering method for underwater robots, leading to the design of soft fish-inspired flippers and tails \cite{white2021tunabot}, flagella \cite{bujard2021resonant}, manta-like fins \cite{liu2022manta} and snakes \cite{kelasidi2016innovation}. One such problem an underwater craft may face is the onset of gusts, i.e. the abrupt change in magnitude and direction of the oncoming flow, causing a sudden deviation from the desired lift of control surfaces. Gusts can create hazardous conditions due to the sudden perturbation to the system. Therefore the problem of autonomously rejecting the effect of gusts on a craft (Gust Load Alleviation, GLA) has been heavily studied in aeronautics \cite{regan2012survey} due to the implications it has on flight safety, especially for smaller crafts which are more vulnerable to disturbances. The problem however, has been less studied for underwater robotics. One approach to addressing gust effects is to rely on observers to estimate the disturbing force on the system and use a control strategy to account for it \cite{sofrony2017flight}. 
\begin{figure}[t]
\centerline{\includegraphics[width=0.83\columnwidth]{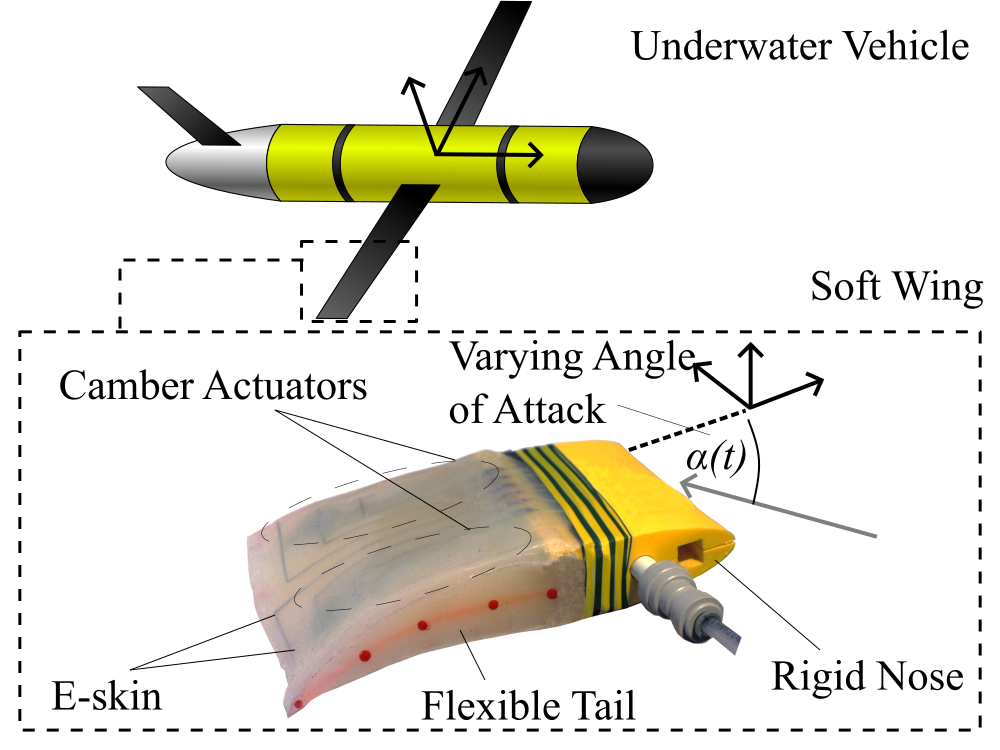}}
\caption{The flexible wing under study. Hydraulic actuators along the length of the wing can be pressurised to increase the curvature. The curvature can be measured using a proprioceptive e-skin.}
\label{coverfigure}
\vspace{-0.3cm}
\end{figure} 

In the case of highly flexible and deformable devices, such as soft robots, the problem of effectively counteracting external environmental disturbances is exacerbated by the still limited availability of suitable soft sensors, \cite{bezawada2022shape}, \cite{ozel2015precise}, \cite{abdelaziz2023state}, which hinder the necessary state estimation capabilities of the systems. On the other hand, the deformable nature of soft robots lends itself to exploiting their structural properties as a mean for exteroceptive sensing in the form of observers. Using proprioceptive information as a means to interpret external forces resembles strategies observed in birds \cite{Martinez2024_kestrel_hovering} and fish \cite{Gibbs2024_station_holding_trouts}.

Here we propose a new gust disturbance observer for a soft morphing wing, aiming to demonstrate how proprioceptive information about the wing's deformation can be leveraged to accurately track and reject the disturbance. To do this, we employ the underwater soft morphing wing from \cite{micklem2022energy}: this soft wing employs soft hydraulic actuators to vary its own curvature,  Fig. \ref{coverfigure}, and a capacitive e-skin, \cite{hu2023stretchable}, to accurately measure such curvature underwater, thus enabling closed-loop control in the absence of disturbances \cite{micklem2024closed}. We aim to demonstrate how a desired lift on a soft wing can be retained in the face of a disturbance which would otherwise cause drastic deviation to the lift. This contribution is achieved as follows:
\begin{itemize}
    \item We develop a dynamic model of a soft morphing wing \cite{micklem2022energy} combining Piecewise Constant Curvature framework \cite{della2023model} with Thin Airfoil Theory, thus enabling us to exploit the well established methodologies in soft robots control;
    \item The model is validated against experimental data from dynamic conditions associated to external hydrodynamic loading, demonstrating its accuracy;
    \item We develop an observer based on an Extended Kalman Filter to demonstrate the capability to estimate abrupt changes in the flow angle of attack (i.e. a gust) from wing camber, testing it in simulation.
    \item Using the knowledge from the observer, we produce a baseline control law to reduce the effect of changes in the angle of attack on the lift.
\end{itemize}
The results demonstrate that, by exploiting the soft, deformable nature of the wing as a proprioceptive sensor, the observer can infer highly dynamic gust disturbances directly from its deformation, paving the way for closed-loop disturbance rejection in aquatic soft robots.

\section{Modelling the underwater soft wing}

We model the underwater soft morphing wing using the Piecewise Constant Curvature (PCC) approximation \cite{della2023model}, as the wing’s kinematics naturally align with this representation (see evidence below) and due to its widespread use in soft robot control. While more accurate beam models, such as the Cosserat model \cite{cosserat1909theorie}, capture the continuum dynamics in greater detail, they make the formulation of the systems' dynamics into state-space form less tractable.

To validate our model of the wing, we use time series from 10 experiments from \cite{micklem2024closed}, where the curvature and lift of the wing are measured in response to changes in pressure of the internal hydraulic actuators and angle of attack while subjected to a steady water flow at $0.2$ ms$^{-1}$. The experimental setup is shown in Fig. \ref{experimental} where the setup was placed in a flume and rotated to different angles of attack.

\subsection{Wing Geometry}

\begin{figure}[t]
\centering

\begin{subfigure}{0.14\textwidth}
\includegraphics[width=\linewidth]{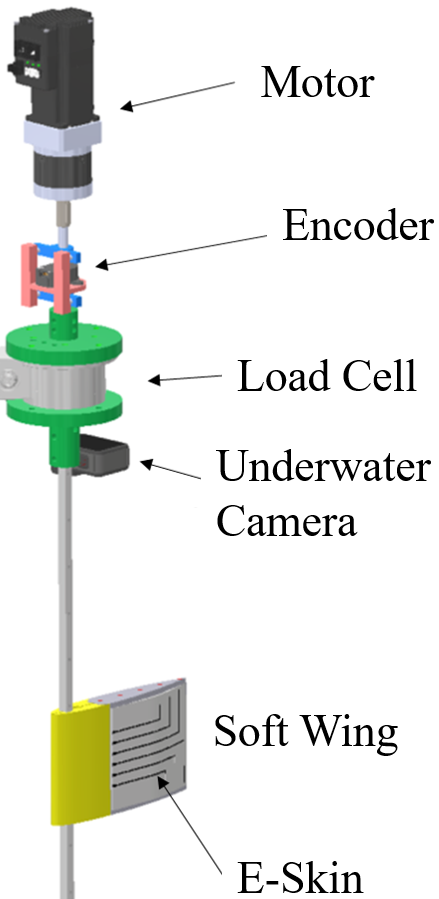}
\caption{ }
\label{experimental}
\end{subfigure}%
\begin{subfigure}{0.30\textwidth}
\includegraphics[width=\linewidth]{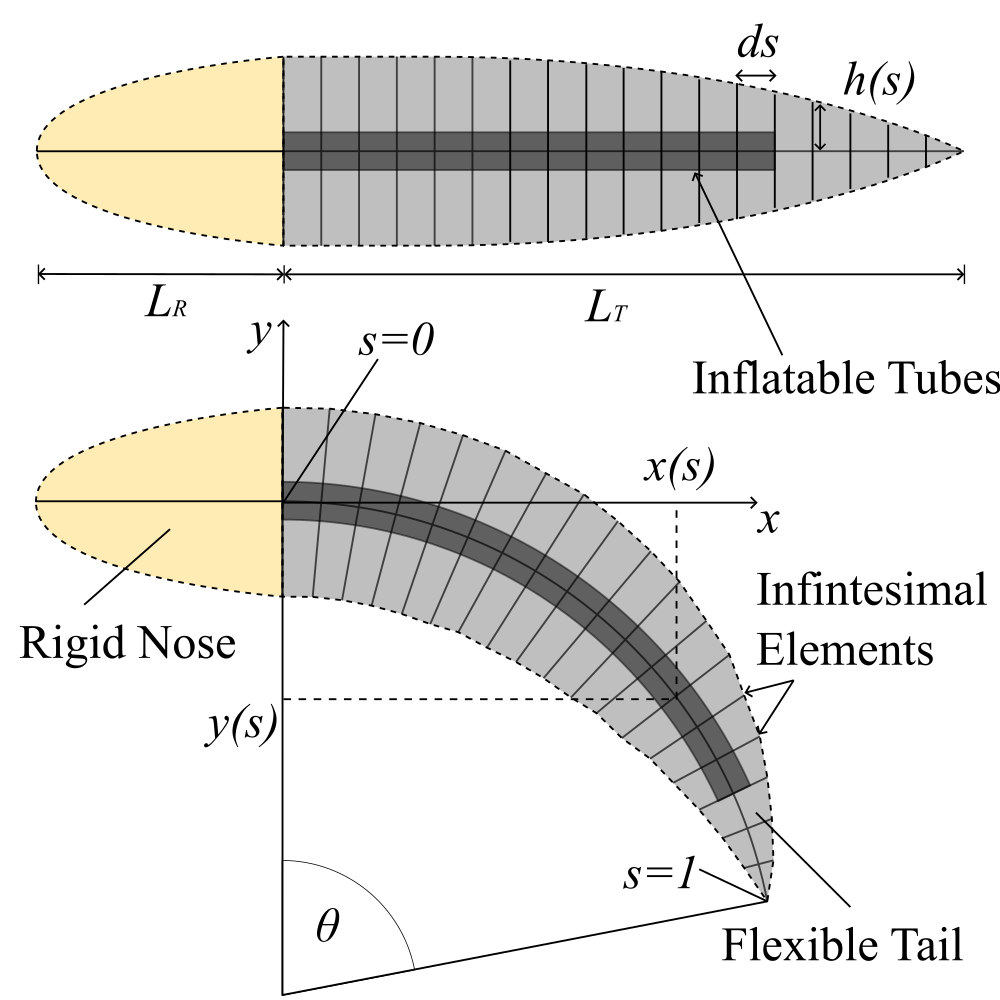}
\caption{ }
\label{geometry}
\end{subfigure}
\caption{a) The experimental setup from \cite{micklem2024closed}. The assembly was placed in a flume. The motor rotated the setup to simulate different angles of attack. The load cell measured lift force. b) The geometry of the wing model. The kinematics of the wing are described in a Cartesian coordinate plane. The parameter $\theta$ describes the curvature of the wing. $s$ describes the normalised distance along the centreline of the wing, with $s=0$ at the base and $s=1$ at the tip. $h(s)$ describes the thickness of the wing.}
\label{geometryandexperimental}
\vspace{-0.3cm}
\end{figure}

The wing has a rigid plastic nose, and a silicon (Ecoflex 00-30) flexible tail. The hydraulic actuators consist of two tubes with natural curvature which run through the silicone, and can be inflated with water to make them stiffer, causing them to bend towards their natural curvature. Varying the pressure allows control to a desired curvature. Fig. \ref{geometry} shows the geometry of the wing. The nose has length $L_R = 0.0785$ m and the tail has length $L_T = 0.1165$ m. The wing also has width $w = 0.12$ m (into the page in Fig. \ref{geometry}). A parameter $s$ describes a point along the wing's centreline, the line along the wings centre running from from the base of the soft section at $s=0$, to the tip at $s=1$. By testing video data from \cite{micklem2024closed} we confirmed the centreline of the wing approximately follows a constant curvature. The wing had 5 visual markers along the flexible section. Arcs were fit to these markers in 2933 image frames, subject to the constraint that the radius of the arc was perpendicular to line between the tip of the nose and the first marker. The average length of the fitted arc was $445.45\pm9.0$ pixels. The average RMSE for the fitted circles across these frames was $4.94\pm0.52$ pixels, which is approximately $0.0013\pm1.36\times10^{-4}$ m, about $1.12\%$ of the wings length. This confirms that the wing's kinematics, and therefore dynamics, can be modelled using the well established PCC approximation. The curvature of the wing is described by the angle $\theta$, which can be mapped from camber measurements from \cite{micklem2024closed}, allowing the use of the experimental data to validate simulations using the PCC approximation.

The kinematics of a point $s$ along the wing are described by Eq. (\ref{forwardkinematics}), where $x(s)$ and $y(s)$ are the Cartesian coordinates and $\theta(s,\theta)$ is the angle of the slice of the wing with respect to the $y$ axis. $J(s,\theta)$ is the Jacobian, given in Eq. (\ref{Jacobian}). 
\begin{equation}
\begin{bmatrix}
x(s, \theta)\\
y(s, \theta)\\
\theta(s, \theta)
\end{bmatrix}=
\begin{bmatrix}
\frac{L_T\sin(s\theta)}{\theta}\\
\frac{L_T(\cos(s\theta)-1)}{\theta}\\
s\theta
\end{bmatrix}
\label{forwardkinematics}
\end{equation}

\begin{equation}
J(s, \theta) =  \begin{bmatrix}
\frac{\partial x(s, \theta)}{\partial \theta}\\
\frac{\partial y(s, \theta)}{\partial \theta}\\
\frac{\partial \theta(s, \theta)}{\partial \theta}
\end{bmatrix}=
\begin{bmatrix}
\frac{L_T(s\theta \cos(s\theta)-\sin(s\theta))}{\theta^2}\\
\frac{L_T(1-\cos(s\theta)-s\theta \sin(s\theta))}{\theta^2}\\
s
\end{bmatrix}
\label{Jacobian}
\end{equation}

The half-thickness of the soft tail $h(s)$ is approximated as
\begin{equation}
    h(s) = -1.362\times10^{-2}s^2 -5.626\times10^{-4}s+1.419\times10^{-2}
    \label{h(s)equation}
\end{equation}
 which was found by measuring points along the wings length and fitting a quadratic function to these points.
 
\subsection{Soft Wing Dynamics}

The dynamics of the wing is formulated making use of the Lagrangian Dynamics of PCC approximation, as with similar approaches previously employed for soft manipulators e.g. \cite{walker2024model}. Following \cite{della2023model}, the dynamics of the wing takes the form,
\begin{equation}
    M(\theta)\ddot{\theta}+C(\theta, \dot{\theta})\dot{\theta} + K(\theta) + K_a(\theta, P) + D\dot{\theta} + \tau_l(\theta, \alpha) = 0
    \label{wingdynamicsform}
\end{equation}
 with mass $M(\theta)$, coriolis $C(\theta,\dot{\theta})$, stiffness $K(\theta)$ and damping $D$ terms typical of a PCC segment as per \cite{della2023model}. In addition, the effect of lift in the configuration space $\tau_l(\theta,\alpha)$ is also modelled. A second, pressure dependent stiffness term is used to model the effect of pressure on the internal rubber tubes $K_a(\theta, P)$, where $P$ is the internal pressure of the tubes, which represents the control parameter since it drives the curvature of the wing. The system is treated as neutrally buoyant, allowing us to neglect the contributions from gravity and buoyancy, as per the experimental setup in \cite{micklem2024closed} (See Fig. \ref{experimental}) where the wing is on its side.
Following \cite{della2023model}, the mass matrix is computed from mass and inertia distributions $m(s)$ and $I(s)$, 
\begin{equation}
    M(\theta) = \int_0^1 J(\theta, s)^\top \begin{bmatrix}
m(s) & 0 & 0\\
0 & m(s) & 0\\
0 & 0 &I(s)
\end{bmatrix} J(\theta, s) ds
\label{massmatrixequation}
\end{equation}
where the mass distribution is given by
\begin{equation}
\begin{split}
m(s) = 2\rho wL_T h(s)
\end{split}
\label{massdistribution}
\end{equation}
obtained by multiplying the wing's rectangular cross section of width $w$ and thickness $2h(s)$  by the density of the silicon $\rho = 1064.6$ kgm$^{-3}$ \cite{Smooth-On} and the length of the tail.
Similarly the inertia distribution $I(s)$ is approximated by considering a slice of the wing at $s$ as a thin rod with length $2h(s)$ and mass $m(s)$:
\begin{equation}
I(s) = \frac{m(s)\left(2h(s)\right)^2}{12}\\
\label{inertiadistribution}
\end{equation}
Finally, the Coriolis term is given as
\begin{equation}
    C(\theta, \dot{\theta})\dot{\theta} = \frac{1}{2}\frac{dM(\theta)}{dt}\dot{\theta} = \frac{1}{2}\frac{\partial M(\theta)}{\partial \theta}\dot{\theta}^2
    \label{coriolisequation}
\end{equation}

\subsection{Stiffness and Damping}

The wing is subject to two stiffness terms which act antagonistically: one is due to the silicone the tail is made from, which applies a restoring force towards the unstrained state of the wing at $\theta=0$; the other is due to the soft hydraulic actuators' rubber tubes, which drive the change in curvature upon pressurization $P$. As such, the latter stiffness term is also a function of the internal pressure $P$ of the tubes. The stiffness of a PCC segment is defined as, 
\begin{equation}
    K(\theta) = \int_0^1 k(s)ds\theta = k\theta
    \label{pccstiffness}
\end{equation}
which describes a torque proportional to the instantaneous curvature $\theta(t)$ and stiffness coefficient $k$, in turn equal to the integral of the local stiffness over the segment.

As the local stiffness $k(s)$ is unknown for the silicon tail, the average stiffness $k$ can instead be estimated by approximating the wing as a beam with constant cross section. Following \cite{gere1997mechanics}, the stiffness for a beam under pure bending is given by,
\begin{equation}
    k = \frac{\lambda I}{L}
    \label{rotationalspring}
\end{equation}
where $\lambda$, $I$ and $L$ are the Youngss modulus, 2nd moment of area and length of the beam respectively. The beam under pure bending will obey $\tau = k\theta$.
Here, the Youngs modulus of Ecoflex 00-30 is $\lambda_w = 100$ kPa \cite{vaicekauskaite2020mapping}, the length of the beam is $L_T$ and the 2nd moment of area $I(s)$ is obtained by averaging that of a beam with rectangular cross-section with width $w$ and height $h(s)$ and then subtracting the contribution from the moment of area of the internal actuators.
The tubes have a circular cross-section with approximate mean radius $r=0.0075$ m, allowing the stiffness of the tail to be given by
\begin{equation}
    k = \frac{\lambda_w I_w}{T_L} = \frac{\lambda_w \left(\frac{w\left(\int_0^12h(s)ds\right)^3}{12} - 2\frac{\pi}{4}r^4\right)}{T_L}
    \label{wingstiffness}
\end{equation}
such that $k$=$0.0834$ Nm rad$^{-1}$.

Next the stiffness of the tubes is considered. The relation between 2nd moment of area of the tubes and pressure $I_t(P)$ 
was estimated experimentally, yielding the following relation,
\begin{equation}
\begin{split}
I_t(P) \approx 4.058\times10^{-23} P^3 -2.030\times10^{-18}
P^2\\ +4.125\times10^{-14}P+2.129\times10^{-9}
\end{split}
    \label{stiffnesspressure}
\end{equation}
Taking the internal tubes length as $L_T$ and Youngs modulus $\lambda_t = 1.7$ MPa (in close agreement with \cite{ashby2012materials} for Butyl rubber), the pressure dependent stiffness of the hydraulic actuators is modelled by substituting $L_T$, $\lambda_t$ and $I_t(P)$ into Eq. (\ref{rotationalspring}),
\begin{equation}
\begin{split}
    k_a(P) = 5.922\times10^{-16} P^3 - 2.962\times10^{-11} P^2\\
    + 6.019\times10^{-7} P + 3.107\times10^{-2}
\end{split}
\label{tubestiffness}
\end{equation}
The second stiffness term is then given by
\begin{equation}
    K_a(\theta,P) = k_a(P)\left(\theta-\frac{\pi}{4}\right)
    \label{actuatingstiffness}
\end{equation}
which dictates that the curvature point of equilibrium is $\theta = \frac{\pi}{4}$ rad due to the wings natural curvature.

Finally, the damping term of a PCC segment is modelled according to \cite{della2023model} as, 
\begin{equation}
    D(\theta)\dot{\theta} = \int_0^1 sd(s)ds\dot{\theta} =\frac{1}{2}d\dot{\theta}
    \label{pccdamping}
\end{equation}
where the local damping coefficient, $d(s)$, has been assumed constant and equivalent to the mean damping coefficient $d=0.1$ Nmsrad$^{-1}$. 

\subsection{Modelling Lift of a Soft Wing}

Definition of a suitable estimation of lift on the PCC section of the soft wing is essential for accurate dynamics modelling and formulation of an appropriate disturbance observer. The model of the lift must be expressed equivalently in Cartesian space as a force $F_l$ (for control of the lift), and in configuration space, as a torque $\tau_l$ for the dynamics. In both cases, the lift is dependent on angle of attack $\alpha$ and curvature $\theta$.

In the absence of sufficient data to construct an accurate fitting function for $F_l$ and $\tau_l$, we resort to Thin Airfoil Theory (TAT), \cite{houghton2003aerodynamics}, as a means to find a function for the lift that spans a sufficient range of $\theta$ and $\alpha$. 
Another approach for modeling fluid forces on a flexible body is Resistive Force Theory (RFT), which has been applied to flagella in \cite{gray1955propulsion}. However, RFT considers only local viscous effects along the body and neglects pressure-induced lift generated by flow around a foil. For this reason, TAT is better suited for capturing the hydrodynamic lift relevant to soft morphing wings
TAT assumes a wing with zero-thickness around its centreline and infinite width-span. These conditions are quite different from the wing used here. Therefore, we employ TAT to produce a baseline function and augment it with a corrective mapping to bring the function in line with the six clusters of equilibrium points from experimental data in Table \ref{avgdata}. To estimate the effects of lift in joint space, we  assume that eq. \ref{wingdynamicsform} is at equilibrium to give,
\begin{equation}
    \tau_l(\theta,\alpha) = -K(\theta)-K_a(\theta,P)
    \label{esttorque}
\end{equation}

\begin{figure}[t]
\centerline{\includegraphics[width=0.8\columnwidth]{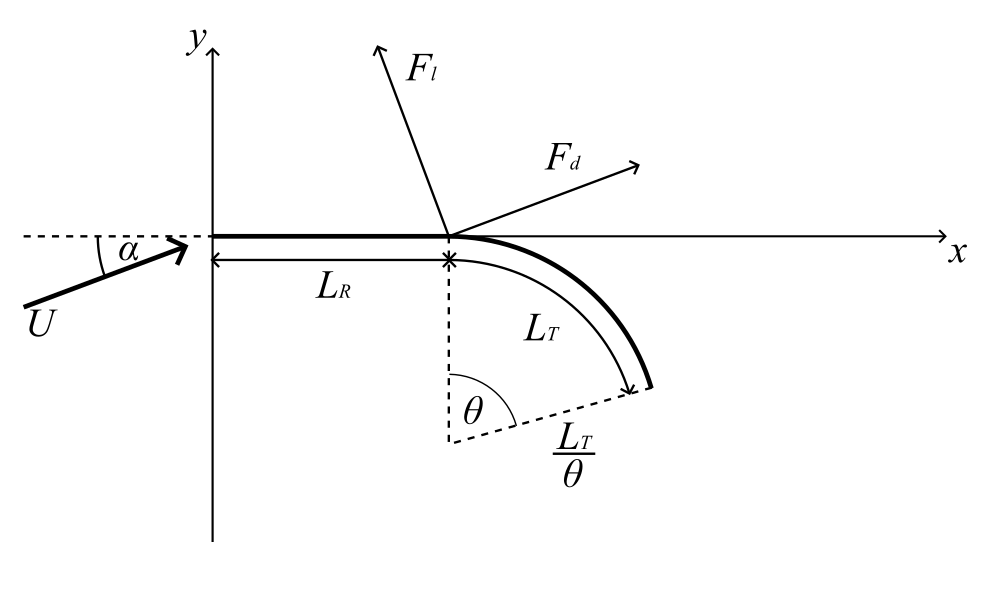}}
\caption{The camberline of the wing, used for thin aerofoil theory. $F_l$ and $F_d$ are the forces from lift and drag, $\alpha$ is the angle of attack and $U$ is the speed of the oncoming flow.}
\label{camberline}
\vspace{-0.3cm}
\end{figure}

TAT formulation requires an expression of the centreline (or \textit{camberline}) of the wing. This is shown schematically in Fig. \ref{camberline} for the case of our wing (accounting for both the flexible tail and the rigid nose), as well as lift and drag forces $F_l$, $F_d$ in the appropriate frame, and the speed of flow $U$ and the angle of attack $\alpha$. $L_R$, $L_T$ and $\theta$ are the previously defined lengths of the rigid and soft parts of the wing and the angle of curvature respectively.

\noindent The equation of the centreline is,
\begin{equation}
    y = \begin{cases}
    0 & 0 \leq x \leq L_R  \\
      \sqrt{\left(\frac{L_t}{\theta}\right)^2-\left(x-L_R\right)^2}-\frac{L_T}{\theta} & L_R < x\leq c
   \end{cases}
   \label{y}
\end{equation}
 where $c$ is the length of the wing along the $x$ axis, given by  
\begin{equation}
    c = L_R + \frac{L_T\sin(\theta)}{\theta};
    \label{c}
\end{equation}
This allows the TAT-based estimate of $F_l$, denoted by $F_{le}$ as, 
\begin{equation}
    F_{le} = w\rho_f U \int_0^c k_l dx = w\rho_f U \int_0^\pi \frac{c}{2} k_l \sin{\beta} d\beta
    \label{l}
\end{equation}
where $w$ is the width of the wing, $\rho_f$ is the density of the fluid (in the case of water, $\rho_f=998$ kgm$^{-3}$ ), $U$ is the flow speed (taken to be $U=0.2$ ms$^{-1}$ from experiments) and $k_l$  is the distribution of vorticity over the wing.
The vorticity distribution $k_l$ can be estimated via TAT as 
\begin{equation}
    k_l = 2U\left((\alpha-A_0) \frac{\cos(\beta)+1}{\sin(\beta)}+\sum_{n=1}^\infty A_n\sin(n\beta)\right) 
    \label{k}
\end{equation}
where the terms $A_0$ and $A_n$ are defined respectively by Eq. (\ref{A0}) and Eq. (\ref{An}) in Appendix I.
A change of variables from $x$ to $\beta$ is normally introduced for the sake of simplification of the algebra,
\begin{equation}
    x = \frac{c}{2}(1-cos(\beta))
    \label{betatransform}
\end{equation}

 Eq. (\ref{l}) can be solved numerically across a range of conditions to find a polynomial estimate of $F_l$.
Evaluation of this estimate against the experimental data of Table \ref{avgdata} showed the expected discrepancy due to divergence from the assumptions of TAT.  
To address this, a corrective linear mapping was fit between the estimated lift $F_{le}$ and the experimentally measured lift $F_l$, resulting in the approximation
\begin{equation}
    F_l(\theta, \alpha) \approx 0.7244\theta+1.126\alpha-0.1407
    \label{flexpanded}
\end{equation} 
which allows the lift in Table \ref{avgdata} to be more accurately predicted.

\begin{table}[t]
\caption{The averages of the clustered data points from experimental data.}
\label{avgdata}
\centering
\begin{tabular}{| c | c | c | c | c |} 
 \hline
 $\theta$ (rad) & $\alpha$ (rad) & Pressure (pa) & $F_l$ (N) & $\tau_l$ (Nm)\\ 
 \hline
 0.3013 & 0 & 7817.6 & 0.1117 & 0.008\\ 
 \hline
 0.2969 & 0.1734 & 1407.7 & 0.2293 & 0.0064\\
 \hline
 0.3036 & 0.2611 & 0 & 0.3473 & 0.0046\\
 \hline
 0.6288 & 0 & 76225 & 0.3078 & -0.0001\\
 \hline
 0.6272 & 0.1734 & 75204 & 0.5042 & -0.0015\\
 \hline
 0.6280 & 0.2601 & 74462 & 0.635 & -0.0032\\
 \hline
\end{tabular}
\end{table}

Next, we look for a suitable mapping from Cartesian-space lift force to configuration-space lift torque. The lumped force $F_l$ cannot be simply mapped to a torque, as it is distributed along the wing, and includes a component from the rigid nose. Instead the distribution of the force along the soft section must be considered to capture dynamic wing deformations. 

To calculate the torque from an arbitrary force distribution $F(s)$ over a PCC segment in configuration space, the following relation is used:
\begin{equation}
    \tau = \int_0^1 J^\top(\theta, s)F(s) ds
    \label{external}
\end{equation}
The force distribution is assumed to be the infinitesimal contributions along $x$ from Eq. (\ref{l}) yielding
\begin{equation}
    F(x) = w\rho Uk_l
    \label{liftforcedistribution}
\end{equation}

The final expression for the lift-dependent torque must also include the drag. By assuming a conservative lift-to-drag ratio of 0.5 for an airfoil analogous to that of \cite{micklem2022energy}, we are able to express the force distribution on the PCC segment after transforming it into the Jacobian reference frame of Eq. \ref{Jacobian} by rotating it w.r.t $\alpha$. This yields
\begin{equation}
    F(x) = \begin{bmatrix}
\cos(\alpha)F_l(x)+\sin(\alpha)|\frac{F_l(x)}{2}|\\
-\sin(\alpha)F_l(x)+\cos(\alpha)|\frac{F_l(x)}{2}|\\
0
\end{bmatrix}
    \label{forcedistribution}
\end{equation}
Then, using 
\begin{equation}
    x = L_R + \frac{L_T\sin(s\theta)}{\theta}
    \label{stransform}
\end{equation}
to perform the appropriate changes in variables for integration (i.e. mapping $k_l$ to be a function of $s$), a numerical solution of the estimated torque $\tau_{le}(\theta,\alpha)$ is found.
Once again the data from Table \ref{avgdata} is used to perform a corrective mapping,
which after expansion and re-fitting yields the following lower order approximate for torque used in the reminder of the work.
\begin{equation}
\tau_l(\theta, \alpha) \approx 10^{-2}\bigl(-2.31\,\theta - 1.61\,\alpha + 1.54\bigr)
\label{tlexpanded}
\end{equation}
\subsection{Actuation Pressure Model}

To conclude the formulation of the wing dynamics, a  model for the internal actuating pressure $P$ is developed. This is needed in order to include a time-delay response between the desired reference input $P_c$ and the transient response of $P$ needed to reach that state. The delay can be approximated by a first order ODE with time constant $\tau_p$
\begin{equation}
    \dot{P}=-\frac{P+P_c}{\tau_p}
    \label{pressure}
\end{equation}
This model represents a response such that pressure will change from 0 to 99\% of $P_c$ in approximately $5\tau_p$, which closely matches the observed experimental behaviour, where the same response requires approximately $5$s, indicating a time-constant $\tau_p=1$s.

\subsection{State Space Model}

For control purposes, it is useful to have a state-space model of the form in $\dot{x}=f(x,u)$ and $y=h(x)$, where $x$ is a vector of state variables, $u$ is a vector of input variables, $f(x,u)$ describes the process dynamics and $h(x)$ describes measurements of the state. By rearranging Eq. (\ref{wingdynamicsform}), and introducing a state vector $x = \left[\theta, \dot{\theta}\right]^{\top}$ where $x_1=\theta$ and $x_2=\dot{\theta}$, the state dynamics can be described by 
\begin{equation}
    \begin{bmatrix}
    \dot{x}_1\\
    \dot{x}_2
    \end{bmatrix}=
    \begin{bmatrix}
    x_2\\
    -\frac{C(x_1, x_2)x_2 + K(x_1) + K_a(x_1, P) + Dx_2 + \tau_l(x_1, \alpha)}{M(x_1)}
    \end{bmatrix}
    \label{state}
\end{equation}
where $u=\left[P, \alpha \right]^{\top}$. 
If required, the pressure dynamics in eq. \ref{pressure} could be included as a third state variable with $x_3=P$ and the input $u = P_c$ instead.
The e-skin provides measurements of $\theta$ so the measurement function is given by $y=h(x)=\theta$. A pressure sensor measures $P$ so it is a known input. $\alpha$ however is unknown.

\subsection{Dynamics Model Validation}

\begin{figure}[t]
\centerline{\includegraphics[width=0.5\textwidth]{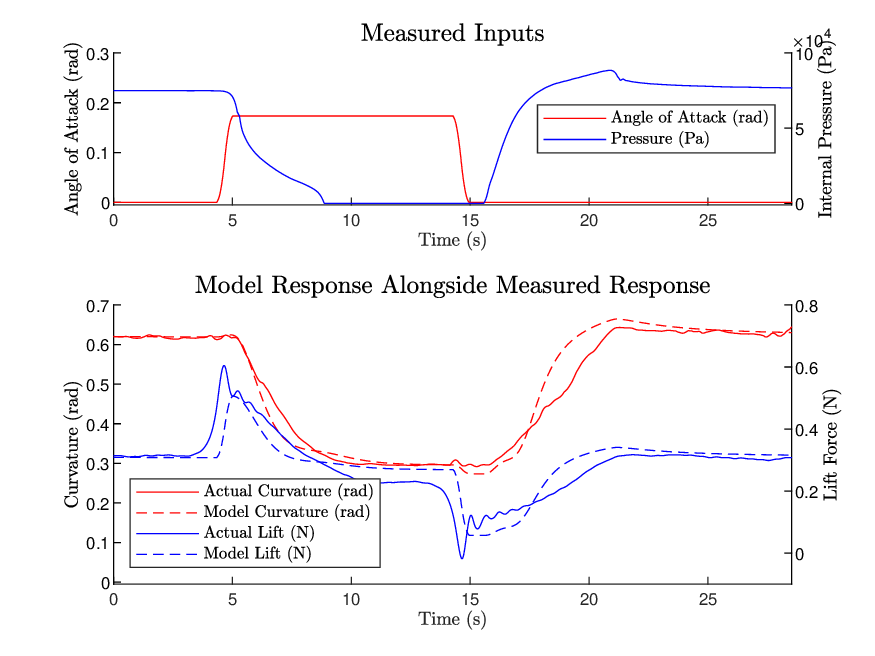}}
\caption{Plot showing the response of the model compared to the real response when subject to the same inputs and initial conditions. The inputs are the internal pressure (manually adjusted), and $\alpha$ stepped from 0 to 0.1745 rad and back.}
\label{val1}
\vspace{-0.3cm}
\end{figure}

\begin{figure}[b]
\centerline{\includegraphics[width=0.5\textwidth]{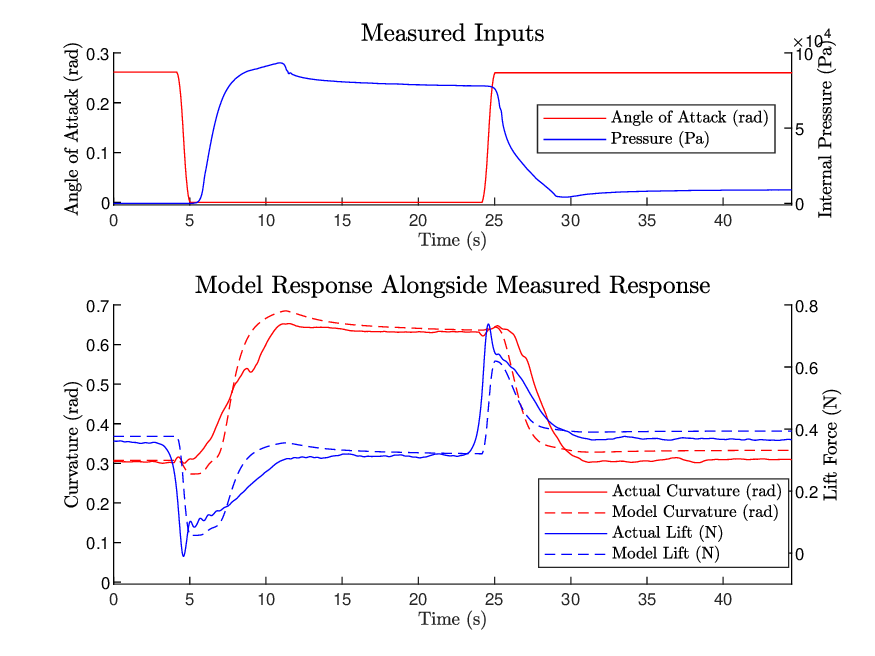}}
\caption{Plot showing the response of the model compared to the real response when subject to the same inputs and initial conditions. The inputs are the internal pressure (manually adjusted), and $\alpha$ stepped from 0.2618 to 0 rad and back.}
\label{val2}
\end{figure}

\begin{figure}[t]
    \centering
    \includegraphics[width=0.48\textwidth]{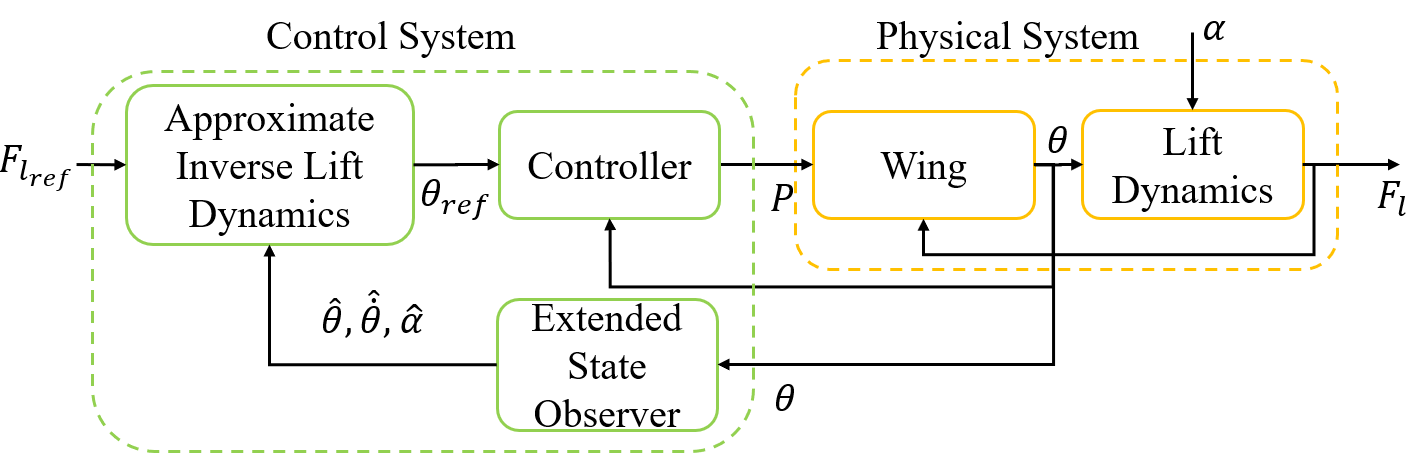}
    \caption{Block diagram showing the proposed control system. The ESO tracks $\alpha$ from measurements of $\theta$. Using a model of the lift a reference curvature $\theta_{ref}$ is produced and tracked by a controller.}
    \label{block}
\end{figure}

To test the validity of the model, measured pressure $P$ and angle of attack $\alpha$ time series data from Micklem \textit{et al.} were used as input to Eq. (\ref{state}) (where pressure dynamics in Eq. (\ref{pressure}) were neglected as $P$ had been measured directly), and the output $\theta$ and the modelled lift $F_l(\theta,\alpha)$ from Eq. (\ref{flexpanded}) were plotted against the corresponding measured time series. Figures \ref{val1} and \ref{val2} depict two sections of interest in these experiments that confirm that the model accurately captures the important features of the dynamics, for the first time demonstrating that PCC can effectively be used as a reliable representation for the dynamics of some flexible wings.

Across the ten series of experiment data, each lasting approximately 80 seconds, the RMSE between predicted and modelled $\theta$ was $0.0339$ rad, and the Mean Absolute Error (MAE) was $0.0222\pm0.0256$ rad. This is $\sim$$1$-$2^{\circ}$ for a system that operates between $\sim$$17$-$35^{\circ}$. This value is relatively low, particularly for a model whose parameters have not been fine-tuned. For lift, the RMSE was $0.0515$ N, with MAE $0.0384\pm0.0344$ N. 
From the figures it can be seen that the model predicts the approximate shapes of the curvature and lift trajectories, and with fine-tuning could be further increased in accuracy. For the purposes of simulation, observer design and control could be made semi-parametric with a method such as \cite{smith2020online}, using a small neural network to account for any un-modelled dynamics. For the purposes of this paper, we assume the parametric model derived so far is sufficient.

\section{Control System}

Using the model, we design and test a controller to reject disturbances in the lift in simulation. Fig. \ref{block} outlines the proposed system. An Extended State Observer (ESO), which includes the angle of attack in the state will estimate the angle of attack from e-skin curvature measurements. By inverting the lift model at the estimated angle of attack, a reference curvature $\theta_{ref}$ to produce a reference lift $F_{l_{ref}}$ can be found. A closed loop controller can then track the desired curvature. This scheme resembles the Disturbance Observer Based Control analysed by Chen \cite{chen2004disturbance}, where the observer and controller can be designed independently to meet different performance criteria.

\subsection{Disturbance Observer}

In order to estimate the $\alpha$ with an observer, the dynamics must be represented as a state space system in which $\alpha$ is include as a state. In order to do this, an extended state vector $x_o = [\theta, \dot{\theta}, \alpha]$ is introduced, which includes $\alpha$. Then $x_{o1} = \theta$, $x_{o2} = \dot{\theta}$ and $x_{o3} = \alpha$. By assuming that $\alpha$ remains mostly constant, subject to ocassional changes, then the dynamics of $\alpha$ can be approximated by $\dot{x}_{o3}=0$. Combining this with Eq. (\ref{state}), the process dynamics $\dot{x}_o=f_o(x_o,u)$ (where $u = P$) of this extended system is given by,
\begin{equation}
\begin{aligned}
\begin{bmatrix}
    \dot{x}_{o1}\\
    \dot{x}_{o2}\\
    \dot{x}_{o3}
    \end{bmatrix}\!=\!
     &= 
\begin{bmatrix}
x_{o2} \\
-\frac{F(x_{o1},x_{o2},P) + \tau_l(x_{o1},\alpha)}{M(x_{o1})} \\
0
\end{bmatrix}\\
\end{aligned}
\label{extendedstate}
\end{equation}
with
\begin{equation*}
\begin{split}
F(x_{o1},x_{o2},P) = {} & C(x_{o1},x_{o2}) x_{o2} + K(x_{o1}) \\
                        & + K_a(x_{o1},P) + D x_{o2}
\end{split}
\end{equation*}
and where the measurements function is $h_o(x_o)=\theta$.

The Extended Kalman Filter (EKF) was chosen for ESO design \cite{simon2006optimal} with the observer system described by 
\begin{equation}
    \hat{\dot{x}}_o = f_o(\hat{x}_o, u) + K_o(\theta-\hat{\theta})
    \label{ekf}
\end{equation}
where the circumflex (i.e. hat) symbol denotes the estimate of a variable. $K_o$ is a matrix of gains, given in terms of $P$, $C$ and $R$, 
\begin{equation}
    K_o = PC^{\top}R^{-1}
    \label{K}
\end{equation}
where $P$ evolves according to 
\begin{equation}
    \dot{P} = AP+PA^{\top}-K_oCP+Q
    \label{P}
\end{equation}

$C$ and $A$ are the Jacobians of $h_o$ and $f_0$ respectively, given by $\frac{dh_o}{dx}$ and $\frac{df_o}{dx}$. $Q$ and $R$ are the covariance matrices of possible additive Gaussian white noise signals $w(t)$ and $v(t)$ to the process, $f_o(x,u)+w(t)$ and input, $h_o(x)+v(t)$. Their values can be tuned to affect response speed and accuracy. These were manually tuned to be $R=1\times10^{-6}$ and 
\begin{equation}
    Q=\begin{bmatrix}
        1\times10^{-2} & 0 & 0\\
        0 & 1\times10^{-2} & 0\\
        0 & 0 & 1
    \end{bmatrix}
\end{equation}

Note that many different observer designs are possible, to satisfy different requirements such as stability, noise rejection, convergence speed, computational simplicity etc. We choose to use the EKF due to its relative simplicity in implementation and tuning and its computational efficiency. However a Lyapunov Stable observer such as that suggested by Chen \cite{chen2004disturbance} or an optimal Moving Horizon Estimator (MHE) (See \cite{rawlings2020model}) could also be selected, depending on performance requirements.

\begin{figure}[t]
\centerline{\includegraphics[width=0.5\textwidth]{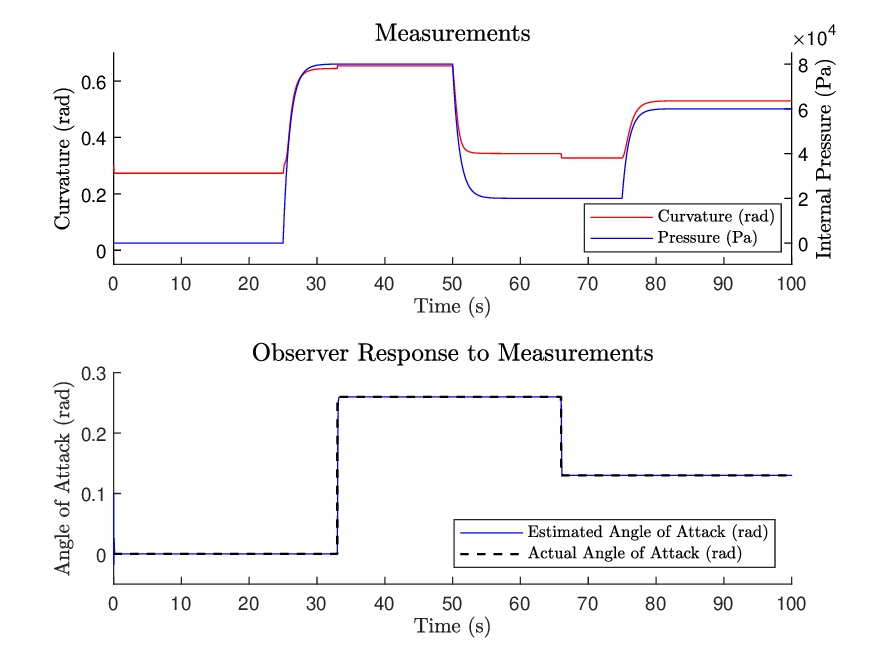}}
\caption{The response of the EKF to a set of test input pressures and angles of attack. Pressure and $\alpha$ were set to a variety of references. Curvature and internal pressure were measured (Top plot) and used as inputs to the EKF. The EKF was able to estimate $\alpha$ when suddenly changing.}
\label{ekfresponse}
\vspace{-0.3cm}
\end{figure}

To test the EKF, the model described by Eq. (\ref{state}) was subjected to a changing angle of attack $\alpha$. A changing pressure input was also used, modelled with Eq. (\ref{pressure}), to demonstrate the observer works across a range of possible curvatures the wing might take. The pressure and $\alpha$ values were chosen to cover the range of pressures and angles of attacks in the experimental data. The system was initialised with $\theta = 0.279$ rad with an initial internal pressure of 0 kPa. The pressure reference was varied between 0, 80, 20 and 60 kPa, while the $\alpha$ was varied between 0, 0.26 and 0.13 rad. The observer was initialised with an incorrect $\theta = 0.3$ rad and $\hat{\alpha}=0.1$. The response of the observer is plotted in Fig. \ref{ekfresponse}.

It can be seen that the observer produces an estimate of $\alpha$ that asymptotically tracks the actual value. Effectively, the observer is using the deformation of the soft wing caused by gusts to estimate the disturbances acting upon it, which, to the best of the authors knowledge is the first time such an idea has been implemented.

\subsection{Disturbance Rejection}

With the observer complete, we design a controller. First, to produce a reference curvature $\theta_{ref}$ to achieve a reference lift $F_{l_{ref}}$, we invert the lift model from Eq. \ref{flexpanded} and substitute $\alpha$ for the estimated value $\hat{\alpha}$:
\begin{equation}
    \theta_{ref} = \frac{F_{l_{ref}}-1.126\hat{\alpha}+0.1407}{0.7244}
\end{equation}. We then implement a simple PI controller to steer $\theta$ to $\theta_{ref}$ and manually tune the proportional gain $K_P = 5\times10^5$ and integral gain $K_I = 5\times10^5$. As with the observer, it should be noted that other controllers could be implemented to meet requirements such as stability, optimality, robustness etc. We choose a PI controller for its ease to implement and tune.

\begin{figure}[t]
\centerline{\includegraphics[width=0.5\textwidth]{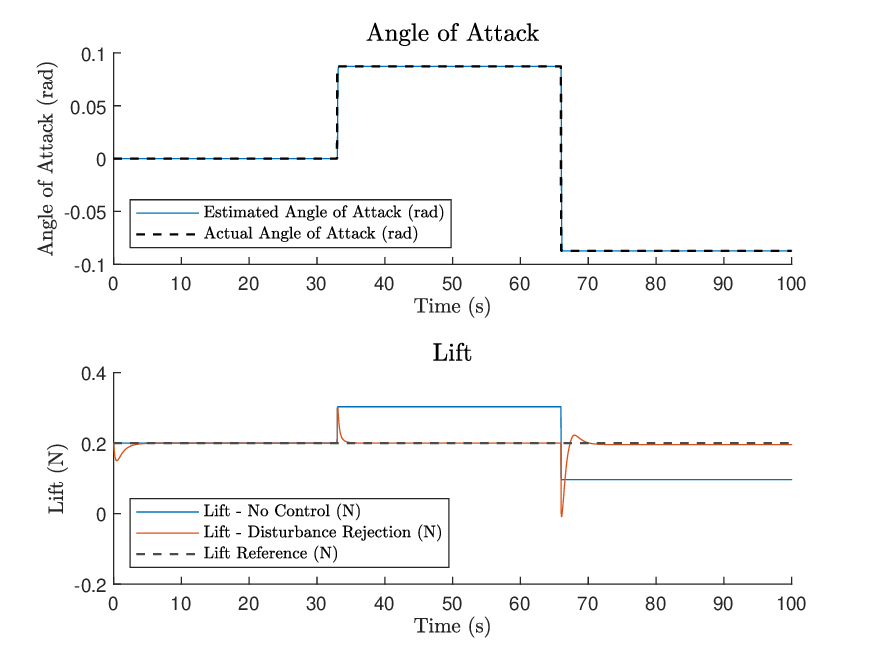}}
\caption{The controller response to step changes in $\alpha$. $\alpha$ is stepped to $\pm5$ degrees (Top) and the observer tracks this change. Then a controller steers the wing to the appropriate curvature to maintain a reference lift (Bottom).}
\label{constantaoa}
\vspace{-0.3cm}
\end{figure}

\begin{figure}[b]
\vspace{-0.3cm}
\centerline{\includegraphics[width=0.5\textwidth]{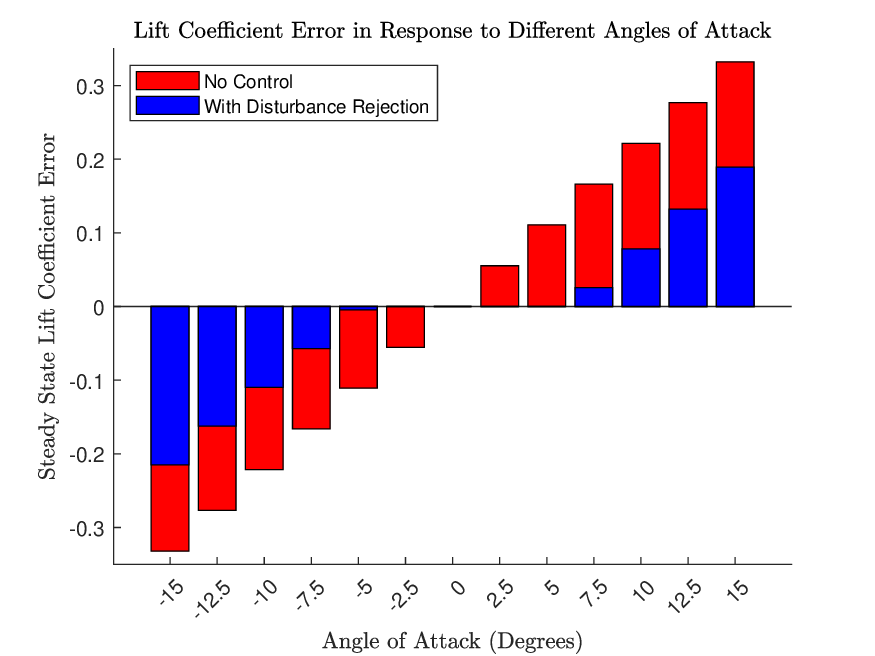}}
\caption{Bar chart showing the error in lift coefficient with and without the controller at steady-state $\alpha$, for a reference lift of $0.2$N.}
\label{liftbar}
\end{figure}

We test the complete control system against step changes to $\alpha$ in Fig. \ref{constantaoa}. $F_{l_{ref}}$ is set to $0.2$N, $\alpha$ is initially $0$ rad. For the wing without control, pressure is set to a constant value $53582$ Pa, known from experiments to yield a curvature that produces $0.2$ N at $\alpha = 0$ rad. We then step $\alpha$ to $\pm0.0872$ rad ($\pm5^{\circ}$). As can be seen, the observer tracks the change and the controller calculates a curvature reference to restore the lift to $0.2$ N. Fig. \ref{liftbar} shows the error in lift coefficient for the wing at a range of $\alpha$ when $F_{l_{ref}}=0.2$ N. It can be seen that the controller completely mitigates the effect of a change to $\alpha$ within a range, after which the wing's curvature saturates. Outside this range, the controller still reduces the effects of the disturbance compared to the constant pressure case.

\begin{figure}[t]
\centerline{\includegraphics[width=0.5\textwidth]{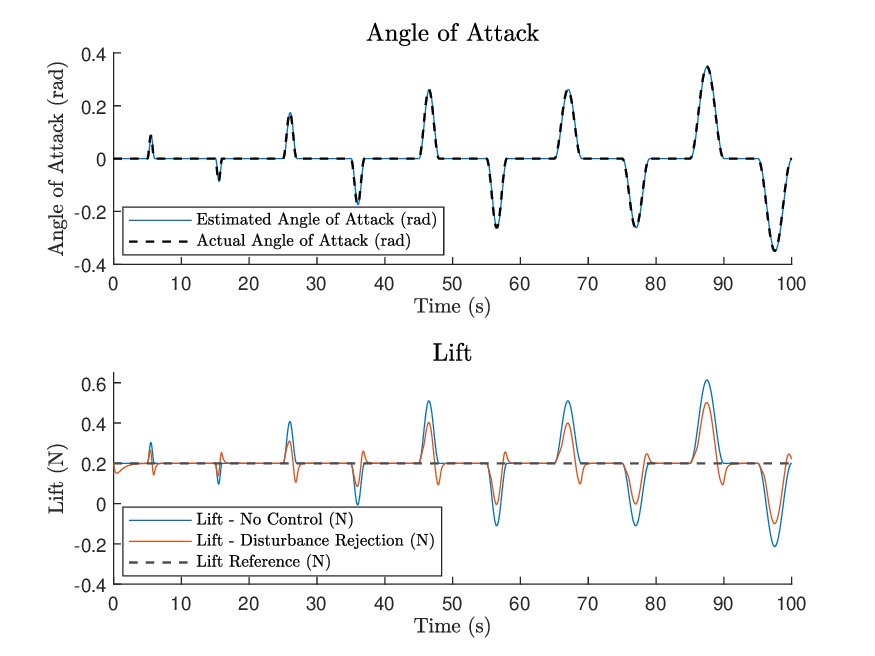}}
\caption{The controller response to 1-cos $\alpha$ changes. $\alpha$ is varied (Top) and the observer tracks these changes. Then a controller steers the wing to the appropriate curvature to maintain a reference lift (Bottom).}
\label{cosaoa}
\vspace{-0.3cm}
\end{figure}

To test a harsher case we subject the wing to several 1-cosine changes to $\alpha$ in Fig. \ref{cosaoa}. The 1-cosine gust models are fairly common e.g. \cite{ghoreyshi2018simulation}, although normally it is used to describe the speed of the flow while here we apply it to describe time-varying $\alpha$. As can be seen from the figure, the case using control reduces the magnitudes of the peaks in lift caused by the disturbances when compared to the constant pressure case. For the lift reference of $0.2$ N, the RMSE for the controlled lift was $0.0734$ N, a significant improvement from the RMSE of $0.1103$ N in the uncontrolled case. Unlike the case of step changes to $\alpha$, the disturbance is not completely rejected due to the bandwidth of the actuators.

\section{Conclusions and Future Work}

In this paper, we developed a model for an underwater soft morphing wing under gust disturbances, integrating the PCC approximation for soft body dynamics with Thin Airfoil Theory to capture hydrodynamic lift. The model shows strong agreement with experimental data and leverages the extensive literature on soft robot control using PCC approximations. By applying an EKF, the observer can estimate parameters describing rapid gust variations, such as changes in angle of attack, enabling the controller to mitigate their effects.

The soft wing’s intrinsic proprioception provides a natural means to sense its own deformation, which, when combined with the observer, allows for effective estimation of external disturbances. While the current study focuses on a single degree of freedom, extending the framework to a 3D wing with multiple deformable modes could enable simultaneous estimation of flow speed and more complex gusts, enhancing overall disturbance rejection.

This work lays the foundation for soft, bio-inspired underwater systems capable of sensing, maneuvering, and propelling themselves effectively in highly perturbed environments, mimicking the remarkable capabilities of living organisms.
\section*{Appendix I: Thin Aerofoil Theory Coefficients}

Equations (\ref{A0}) and (\ref{An}) define coefficients used in TAT.
\begin{equation}
    A_0 = \frac{1}{\pi}\int_0^\pi \frac{dy}{dx} d\beta
    \label{A0}
\end{equation}
\begin{equation}
    A_n = \frac{2}{\pi}\int_0^\pi \frac{dy}{dx} \cos(n\beta) d\beta
    \label{An}
\end{equation}


\bibliographystyle{IEEEtran}
\bibliography{IEEEabrv,ref}

@inproceedings{micklem2022energy,
  title={Energy-efficient tunable-stiffness soft robots using second moment of area actuation},
  author={Micklem, Leo and Weymouth, Gabriel D and Thornton, Blair},
  booktitle={2022 IEEE/RSJ International Conference on Intelligent Robots and Systems (IROS)},
  pages={5464--5469},
  year={2022},
  organization={IEEE}
}

@article{gray1955propulsion,
  title={The propulsion of sea-urchin spermatozoa},
  author={Gray, James and Hancock, Gregory J},
  journal={Journal of Experimental Biology},
  volume={32},
  number={4},
  pages={802--814},
  year={1955},
  publisher={The Company of Biologists Ltd}
}

@article{micklem2024closed,
  title={Closed-loop underwater soft robotic foil shape control using flexible e-skin},
  author={Micklem, Leo and Dong, Huazhi and Giorgio-Serchi, Francesco and Yang, Yunjie and Weymouth, Gabriel D and Thornton, Blair},
  journal={arXiv preprint arXiv:2408.01130},
  year={2024}
}

@article{della2023model,
  title={Model-based control of soft robots: A survey of the state of the art and open challenges},
  author={Della Santina, Cosimo and Duriez, Christian and Rus, Daniela},
  journal={IEEE Control Systems Magazine},
  volume={43},
  number={3},
  pages={30--65},
  year={2023},
  publisher={IEEE}
}

@inproceedings{walker2024model,
  title={Model predictive wave disturbance rejection for underwater soft robotic manipulators},
  author={Walker, Kyle L and Della Santina, Cosimo and Giorgio-Serchi, Francesco},
  booktitle={2024 IEEE 7th International Conference on Soft Robotics (RoboSoft)},
  pages={40--47},
  year={2024},
  organization={IEEE}
}

@book{simon2006optimal,
  title={Optimal state estimation: Kalman, H infinity, and nonlinear approaches},
  author={Simon, Dan},
  year={2006},
  publisher={John Wiley \& Sons}
}

@article{gere1997mechanics,
  title={Mechanics of Materials, ed},
  author={Gere, James M and Timoshenko, SP},
  journal={Boston, MA: PWS},
  year={1997}
}

@article{vaicekauskaite2020mapping,
  title={Mapping the mechanical and electrical properties of commercial silicone elastomer formulations for stretchable transducers},
  author={Vaicekauskaite, Justina and Mazurek, Piotr and Vudayagiri, Sindhu and Skov, Anne Ladegaard},
  journal={Journal of Materials Chemistry C},
  volume={8},
  number={4},
  pages={1273--1279},
  year={2020},
  publisher={Royal Society of Chemistry}
}

@book{ashby2012materials,
  title={Materials and the environment: eco-informed material choice},
  author={Ashby, Michael F},
  year={2012},
  publisher={Elsevier}
}

@book{houghton2003aerodynamics,
  title={Aerodynamics for engineering students},
  author={Houghton, Edward Lewis and Carpenter, Peter William},
  year={2003},
  publisher={Elsevier}
}

@article{hughes2016soft,
  title={Soft manipulators and grippers: A review},
  author={Hughes, Josie and Culha, Utku and Giardina, Fabio and Guenther, Fabian and Rosendo, Andre and Iida, Fumiya},
  journal={Frontiers in Robotics and AI},
  volume={3},
  pages={69},
  year={2016},
  publisher={Frontiers Media SA}
}

@article{shintake2018soft,
  title={Soft robotic grippers},
  author={Shintake, Jun and Cacucciolo, Vito and Floreano, Dario and Shea, Herbert},
  journal={Advanced materials},
  volume={30},
  number={29},
  pages={1707035},
  year={2018},
  publisher={Wiley Online Library}
}

@article{rich2018untethered,
  title={Untethered soft robotics},
  author={Rich, Steven I and Wood, Robert J and Majidi, Carmel},
  journal={Nature Electronics},
  volume={1},
  number={2},
  pages={102--112},
  year={2018},
  publisher={Nature Publishing Group UK London}
}

@article{bujard2021resonant,
  title={A resonant squid-inspired robot unlocks biological propulsive efficiency},
  author={Bujard, Thierry and Giorgio-Serchi, Francesco and Weymouth, Gabriel D},
  journal={Science Robotics},
  volume={6},
  number={50},
  pages={eabd2971},
  year={2021},
  publisher={American Association for the Advancement of Science}
}

@article{hu2023stretchable,
  title={Stretchable e-skin and transformer enable high-resolution morphological reconstruction for soft robots},
  author={Hu, Delin and Giorgio-Serchi, Francesco and Zhang, Shiming and Yang, Yunjie},
  journal={nature machine intelligence},
  volume={5},
  number={3},
  pages={261--272},
  year={2023},
  publisher={Nature Publishing Group UK London}
}

@inproceedings{sofrony2017flight,
  title={Flight control system design for wind gust rejection based on an Unknown Input Observer and a Simple Adaptive Controller},
  author={Sofrony, Jorge and Turner, Matthew C},
  booktitle={2017 IEEE Conference on Control Technology and Applications (CCTA)},
  pages={1961--1966},
  year={2017},
  organization={IEEE}
}

@article{liu2022manta,
  title={A manta ray robot with soft material based flapping wing},
  author={Liu, Qimeng and Chen, Hao and Wang, Zhenhua and He, Qu and Chen, Linke and Li, Weikun and Li, Ruipeng and Cui, Weicheng},
  journal={Journal of Marine Science and Engineering},
  volume={10},
  number={7},
  pages={962},
  year={2022},
  publisher={MDPI}
}

@manual{Smooth-On,
    url={\url{https://www.smooth-on.com/tb/files/ECOFLEX_SERIES_TB.pdf}},
    title={Ecoflex™ 00-30},
    organisation={Smooth-On},
    note = {[Accessed 27-02-2025]}
}

@techreport{regan2012survey,
  title={Survey of applications of active control technology for gust alleviation and new challenges for lighter-weight aircraft},
  author={Regan, Christopher D and Jutte, Christine V},
  year={2012},
  institution={NASA},
}

@article{white2021tunabot,
  title={Tunabot Flex: A tuna-inspired robot with body flexibility improves high-performance swimming},
  author={White, Carl H and Lauder, George V and Bart-Smith, Hilary},
  journal={Bioinspiration \& Biomimetics},
  volume={16},
  number={2},
  pages={026019},
  year={2021},
  publisher={IOP Publishing}
}

@article{kelasidi2016innovation,
  title={Innovation in underwater robots: Biologically inspired swimming snake robots},
  author={Kelasidi, Eleni and Liljeback, Pal and Pettersen, Kristin Y and Gravdahl, Jan Tommy},
  journal={IEEE robotics \& automation magazine},
  volume={23},
  number={1},
  pages={44--62},
  year={2016},
  publisher={IEEE}
}

@article{bezawada2022shape,
  title={Shape Reconstruction of Soft Manipulators Using Vision and IMU Feedback},
  author={Bezawada, Harish and Woods, Cole and Vikas, Vishesh},
  journal={IEEE Robotics and Automation Letters},
  volume={7},
  number={4},
  pages={9589--9596},
  year={2022},
  publisher={IEEE}
}

@article{ozel2015precise,
  title={A precise embedded curvature sensor module for soft-bodied robots},
  author={Ozel, Selim and Keskin, Nehir A and Khea, Darien and Onal, Cagdas D},
  journal={Sensors and Actuators A: Physical},
  volume={236},
  pages={349--356},
  year={2015},
  publisher={Elsevier}
}

@article{abdelaziz2023state,
  title={State Estimation of Continuum Robots: A Nonlinear Constrained Moving Horizon Approach},
  author={Abdelaziz, Hend and Nada, Ayman and Ishii, Hiroyuki and El-Hussieny, Haitham},
  journal={arXiv preprint arXiv:2308.03931},
  year={2023}
}

@article{Martinez2024_kestrel_hovering,
    author = {Martinez Groves-Raines, Mario and Yi, George and Penn, Matthew and Watkins, Simon and Windsor, Shane and Mohamed, Abdulghani},
    title = {Steady as they hover: kinematics of kestrel wing and tail morphing during hovering flights},
    journal = {Journal of Experimental Biology},
    volume = {227},
    number = {15},
    pages = {jeb247305},
    year = {2024},
    month = {08},
    issn = {0022-0949},
    doi = {10.1242/jeb.247305},
    url = {https://doi.org/10.1242/jeb.247305},
    eprint = {https://journals.biologists.com/jeb/article-pdf/227/15/jeb247305/3558608/jeb247305.pdf},
}

@book{cosserat1909theorie,
  title={Th{\'e}orie des corps d{\'e}formables},
  author={Cosserat, Eug{\`e}ne Maurice Pierre and Cosserat, Fran{\c{c}}ois},
  year={1909},
  publisher={A. Hermann et fils}
}

@article{smith2020online,
  title={Online simultaneous semi-parametric dynamics model learning},
  author={Smith, Joshua and Mistry, Michael},
  journal={IEEE Robotics and Automation Letters},
  volume={5},
  number={2},
  pages={2039--2046},
  year={2020},
  publisher={IEEE}
}

@article{chen2004disturbance,
  title={Disturbance observer based control for nonlinear systems},
  author={Chen, Wen-Hua},
  journal={IEEE/ASME transactions on mechatronics},
  volume={9},
  number={4},
  pages={706--710},
  year={2004},
  publisher={IEEE}
}

@book{rawlings2020model,
  title={Model predictive control: theory, computation, and design},
  author={Rawlings, James Blake and Mayne, David Q and Diehl, Moritz and others},
  volume={2},
  year={2020},
  publisher={Nob Hill Publishing Madison, WI}
}

@article{ghoreyshi2018simulation,
  title={Simulation and modeling of rigid aircraft aerodynamic responses to arbitrary gust distributions},
  author={Ghoreyshi, Mehdi and Greisz, Ivan and Jirasek, Adam and Satchell, Matthew},
  journal={Aerospace},
  volume={5},
  number={2},
  pages={43},
  year={2018},
  publisher={MDPI}
}

@article{Gibbs2024_station_holding_trouts,
    author = {Gibbs, Brendan J. and Akanyeti, Otar and Liao, James C.},
    title = {Kinematics and muscle activity of pectoral fins in rainbow trout (Oncorhynchus mykiss) station holding in turbulent flow},
    journal = {Journal of Experimental Biology},
    volume = {227},
    number = {5},
    pages = {jeb246275},
    year = {2024},
    month = {03},
    issn = {0022-0949},
    doi = {10.1242/jeb.246275},
    url = {https://doi.org/10.1242/jeb.246275},
    eprint = {https://journals.biologists.com/jeb/article-pdf/227/5/jeb246275/3380012/jeb246275.pdf},
}

\end{document}